\documentclass[runningheads,a4paper]{llncs}

\usepackage{amssymb}
\usepackage{graphicx}

\usepackage{url}
\urldef{\mailsa}\path|{alfred.hofmann, ursula.barth, ingrid.haas, frank.holzwarth,|
\urldef{\mailsb}\path|anna.kramer, leonie.kunz, christine.reiss, nicole.sator,|
\urldef{\mailsc}\path|erika.siebert-cole, peter.strasser, lncs}@springer.com|    
\newcommand{\keywords}[1]{\par\addvspace\baselineskip
\noindent\keywordname\enspace\ignorespaces#1}

\usepackage{tabularx}
\usepackage[subtle]{savetrees}
\usepackage[noadjust]{cite}
\usepackage{authblk}
\usepackage{tablefootnote}
\usepackage{hyperref}

\begin{document}

\mainmatter  

\title{Fast, Simple Calcium Imaging Segmentation \\ with Fully Convolutional Networks}

\titlerunning{Fast, Simple Calcium Imaging Segmentation}

\author{Aleksander Klibisz\inst{1} \and Derek Rose\inst{1} \and Matthew Eicholtz\inst{1} \and \\ \vspace{-0.3cm} Jay Blundon\inst{2} \and Stanislav Zakharenko\inst{2}}

\authorrunning{Klibisz, Rose, Eicholtz, Blundon, Zakharenko}

\tocauthor{Aleksander Klibisz, Derek Rose, Matthew Eicholtz, Jay Blundon, Stanislav Zakharenko}

\institute{Oak Ridge National Laboratory, Oak Ridge, TN 37831 USA \\
\email{\{klibisza,rosedc,eicholtzmr\}@ornl.gov},
\and
St. Jude Children's Research Hospital, Memphis, TN 38105 USA \\
\email{\{jay.blundon,stanislav.zakharenko\}@stjude.org}}


\toctitle{Fast, Simple Calcium Imaging Segmentation}

\maketitle

\begin{abstract}

Calcium imaging is a technique for observing neuron activity as a series of images showing indicator fluorescence over time. Manually segmenting neurons is time-consuming, leading to research on automated calcium imaging segmentation (ACIS). We evaluated several deep learning models for ACIS on the Neurofinder competition datasets and report our best model: U-Net2DS, a fully convolutional network that operates on 2D mean summary images. U-Net2DS requires minimal domain-specific pre/post-processing and parameter adjustment, and predictions are made on full $512\times512$ images at $\approx$9K images per minute. It ranks third in the Neurofinder competition ($F_1=0.57$) and is the best model to exclusively use deep learning. We also demonstrate useful segmentations on data from outside the competition. The model's simplicity, speed, and quality results make it a practical choice for ACIS and a strong baseline for more complex models in the future.

{\let\thefootnote\relax\footnote{This manuscript has been authored by UT-Battelle, LLC under Contract No. DE-AC05-00OR22725 with the U.S. Department of Energy. The United States Government retains and the publisher, by accepting the article for publication, acknowledges that the United States Government retains a non-exclusive, paid-up, irrevocable, worldwide license to publish or reproduce the published form of this manuscript, or allow others to do so, for United States Government purposes. The Department of Energy will provide public access to these results of federally sponsored research in accordance with the DOE Public Access Plan (http://energy.gov/downloads/doe-public-access-plan).}}\par

\vspace{-0.5cm}
\end{abstract}

\section{Introduction}

Calcium imaging recordings show neurons from a lab specimen illuminating over time, ranging from minutes to hours in length with resolution on the order of $512\times512$ pixels. Identifying the neurons is one step in a workflow that typically involves motion correction and peak identification with the goal of understanding the activity of large populations of neurons. Given a 3D spatiotemporal (height, width, time) series of images, an ACIS model produces a 2D binary mask identifying neuron pixels.

Until recently, popular ACIS models were mostly unsupervised and required considerable assumptions about the data when making predictions. For example, \cite{kaifosh2014sima, pachitariu2013extracting, pachitariu2016suite2p, pnevmatikakis2016simultaneous} each require the expected number of neurons and/or their dimensions to segment a new series. Moreover, many models were tested on different datasets, making objective comparison difficult. 

The ongoing Neurofinder competition \cite{neurofinder} has nineteen labeled training datasets and nine testing datasets. To our knowledge, this is the best benchmark for ACIS, and the quantity of data makes it appealing for deep learning. Our best model for the Neurofinder datasets is U-Net2DS, an adaptation of the fully convolutional U-Net architecture \cite{ronneberger2015u} that takes 2D summary images as input. Compared to other models that do not use deep learning, U-Net2DS requires minimal assumptions, parameter adjustment, and pre/post-processing when making predictions. The fully convolutional architecture enables training on small windows and making predictions on full summary images. It ranks third in the Neurofinder competition and also shows robustness to datasets for which it was not directly trained. The Keras \cite{chollet2015keras} implementation and trained weights are available at \url{https://github.com/alexklibisz/deep-calcium}.

\section{Related Work}

Deep learning has been explored extensively for medical image analysis, covered thoroughly in \cite{litjens2017survey}. Fully convolutional networks with skip connections were developed for semantic segmentation of natural images \cite{long2015fully} and applied to 2D medical images \cite{ronneberger2015u} and 3D volumes \cite{cciccek20163d, milletari2016v}. An analysis of the role of skip connections in fully convolutional architectures for biomedical segmentation is offered by \cite{drozdzal2016importance}.

ACIS models can be grouped into several categories: matrix factorization, clustering, dictionary learning, and deep learning. In general, matrix factorization models (\cite{petersen2017scalpel, pnevmatikakis2016simultaneous, pachitariu2016suite2p, maruyama2014detecting, pnevmatikakis2013sparse, mukamel2009automated}) consider a calcium imaging series as a matrix of spatial and temporal components and propose methods for extracting the signals. Clustering models (\cite{spaen2017hnccorr, kaifosh2014sima}) define similarity measures for pairs of pixels and divide them using partitioning algorithms. Dictionary learning models (\cite{petersen2017scalpel, pachitariu2013extracting}) extract or generate neuron templates and use them to match new neurons.
Finally, deep learning models learn hierarchies of relevant neuron features from labeled datasets. The deep networks proposed by \cite{apthorpe2016automatic} use 2D and 3D convolutions with some pre/post-processing, and \cite{conv2d} uses 2D convolutions. Both use fully-connected classification layers with fixed-size sliding window inputs and outputs.

\section{Data and Metrics}

\subsection{Calcium Imaging Datasets}

The Neurofinder \cite{neurofinder} training datasets contain 60K images with 7K neurons labeled by researchers at four labs with various experimental settings. Testing labels are withheld and submissions are evaluated on the competition server. Dataset samples are shown in Fig. \ref{fig:dataset_samples} and Fig. \ref{fig:neurofinder_predictions}. To complement this data, we included our own calcium imaging datasets, referred to as the St. Jude datasets. They contain GCaMP6f-expressing neurons from the auditory cortex, a cortical region not represented in Neurofinder, and have bounding box ground truth masks instead of precise outlines. Neurofinder and St. Jude datasets are motion-corrected, 16-bit TIFFs with $512\times512$ resolution and length ranging from 1800 to 8000 images.

We found several noteworthy challenges in the Neurofinder datasets. They are highly heterogeneous in appearance, making it difficult to define a characteristic neuron. Images have highly variable brightness with mean pixel values ranging from 57 to 2998 for individual datasets. The labels have a strong imbalance with an average of 12\% of pixels labeled as neurons. Finally, labeling preference leaves the possibility for inconsistent ground truth. For example, the 04.00 labels were optimized for a particular type of experiment and might be considered inconsistent relative to other datasets.\footnote{See discussion: \url{https://github.com/codeneuro/neurofinder/issues/25}}

\subsection{Summary Images}

\begin{figure}[!t]
    \centering
    \includegraphics[trim={3.5cm 2.5cm 3.0cm 1cm}, clip,width=0.98\columnwidth]{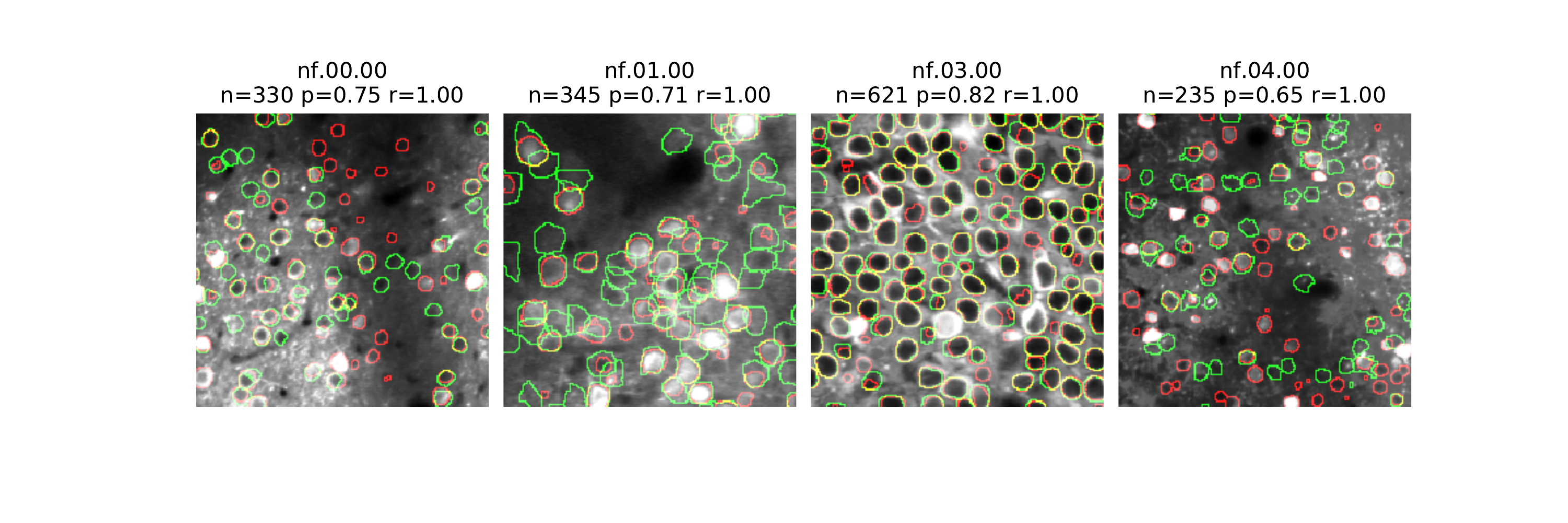}
    \caption{$200\times200$ cropped mean summary images from four Neurofinder training datasets. Ground truth and predicted neurons are outlined in green and red, respectively. Titles denote the ground truth number of neurons and precision and recall metrics for predictions.}
    \label{fig:dataset_samples}
\end{figure}

A common mode of visualization for calcium imaging is a 2D summary image, created by applying a function to each pixel across all frames to flatten the time dimension. Similar to \cite{conv2d, apthorpe2016automatic}, we found mean summary images produced a clear picture of most neurons with low computational cost. Examples are shown in Fig. \ref{fig:dataset_samples} and Fig. \ref{fig:neurofinder_predictions}. Other summary functions include the maximum, minimum, and standard deviation. 

\subsection{Evaluation Metrics}

The Neurofinder competition measures $F_1$, precision, recall, inclusion, and exclusion, with submissions ranked by the mean $F_1$ score across all test datasets. Precision and recall are computed by matching each ground truth neuron to the spatially nearest predicted neuron without replacement. In contrast to the same pixelwise metrics, one predicted neuron may only correspond to a single ground truth neuron, even if the predicted neuron's region encompasses multiple ground truth neurons. Inclusion and exclusion quantify the structural similarity of matched ground truth and predicted neurons.\footnote{Details and implementation: \url{https://github.com/codeneuro/neurofinder-python}}

\section{Proposed Models}

\subsection{U-Net with 2D Summary Images}

\subsubsection{Architecture}

\begin{figure}[!t]
    \centering
    \includegraphics[trim={0cm 0.35cm 0cm 0cm}, width=0.99\columnwidth]{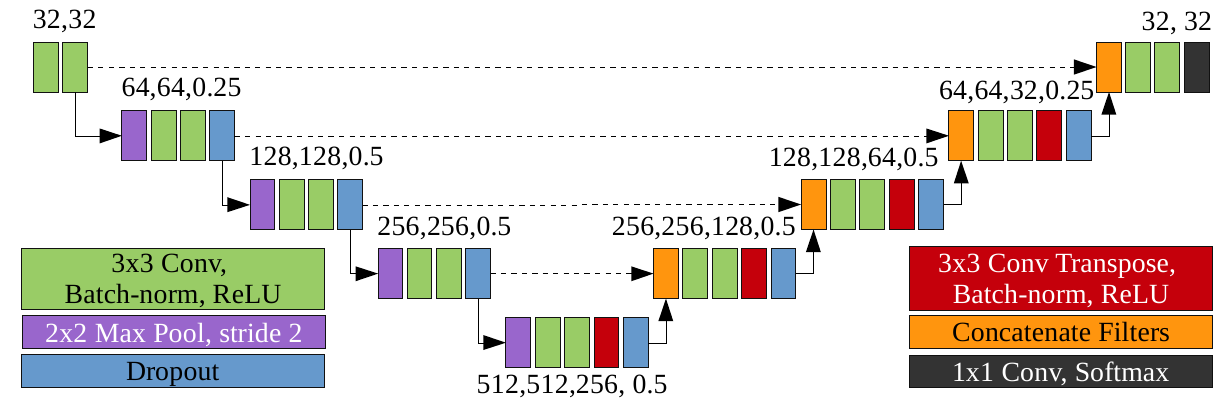}
    \caption{U-Net2DS architecture. Numbers alongside blocks of layers indicate the number of convolutional filters,  convolutional-transpose filters, and dropout probability.}
    \label{fig:U-Net2ds_architecture}
\end{figure}

The fully convolutional U-Net architecture \cite{ronneberger2015u} has four convolution / max-pooling blocks that reduce dimensionality followed by four convolution / transpose-convolution (deconvolution) blocks to restore input dimensionality. A ReLU activation follows each convolutional layer and skip-connections pass the output from each convolution / max-pooling block to corresponding convolution / transpose-convolution blocks via filter concatenation. U-Net2DS makes small modifications to the original U-Net to reduce overfitting. We added zero padding to each convolutional layer, reduced the number of filters at each convolutional layer by 50\%, added batch normalization \cite{ioffe2015batch} after each convolutional layer, and increased dropout. The final layers are a $1\times1$ convolution with two filters followed by a softmax activation, producing a $h\times w \times 2$-channel mask, where the two channels are the probabilities of each pixel belonging to a neuron or background. U-Net2DS has 7.8M parameters and is illustrated in Fig. \ref{fig:U-Net2ds_architecture}.

\subsubsection{Training}

We trained a single U-Net2DS on all Neurofinder training datasets. Each summary image was normalized by subtracting its mean and dividing by its standard deviation. For each dataset, we combined all of the neuron masks into a single mask and removed overlapping and adjacent pixels belonging to different neurons to preserve the true number of independent neurons. We used the top 75\% of each summary image for training and bottom 25\% for validation. We sampled random batches of twenty $128\times128$ image and mask windows, ensuring each window contained a neuron, and applied random rotations and flips. We trained for ten epochs with 100 training steps per epoch using the Adam optimizer \cite{kingma2014adam} with a 0.002 learning rate. After each epoch, we made full-image predictions on the validation data and computed the Neurofinder $F_1$ metric. We saved weights when the mean $F_1$ improved. Pixelwise $F_1$ plateaued around 0.7 and Neurofinder $F_1$ around 0.85. Training on an NVIDIA Titan-X GPU took under half an hour. 

\subsubsection{Loss Function}

On average, the $128\times128$ training windows contained 9\% neuron pixels. This imbalance led us to consider the modified Dice coefficient (MDC) loss function \cite{milletari2016v}. We compared MDC to standard logarithmic loss (LL) and found higher pixelwise recall with MDC, but the average test $F_1$ for LL was consistently better. LL computes each pixel's loss independently, which allows weighting for recall or precision. We experimented with this by multiplying false negative pixel losses between $2\times$ and $10\times$ and saw increased pixelwise recall but lower $F_1$. We ultimately found that standard LL resulted in the best $F_1$ scores.

\subsubsection{Prediction}

Full-image prediction was implemented by reshaping U-Net2DS to take $512\times512$ inputs. This change simplified the implementation and eliminated mask tiling artifacts that were present when using sliding windows. We found test-time augmentation consistently improved the mean $F_1$. This consists of averaging predictions for eight rotations and reflections of the summary image. Given raw TIFF files, the summaries and predictions are computed at $\approx$9K images per minute.

\subsubsection{Results on Neurofinder and St. Jude Datasets}

\begin{figure}[!t]
    \centering
    \includegraphics[trim={3.5cm 2.5cm 3.5cm 0.5cm}, width=1\columnwidth]{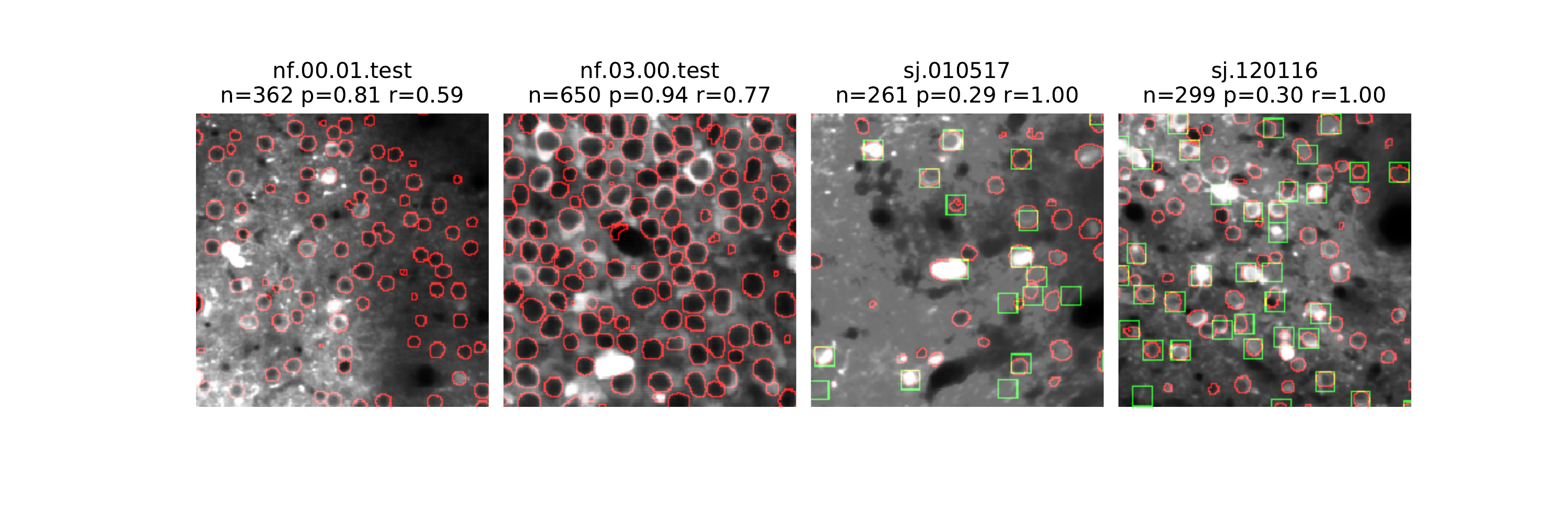}
    \caption{$200\times200$ cropped mean summary images from Neurofinder test and St. Jude datasets with number of predicted neurons, precision, and recall in titles. Ground truth and predicted neurons are outlined in green and red, respectively. Ground truth for Neurofinder test data is not publicly available.}
    \label{fig:neurofinder_predictions}
\end{figure}

\begin{figure}[!t]
    \centering
    \includegraphics[trim={2.5cm 1cm 2.5cm 0.5cm}, width=1\columnwidth]{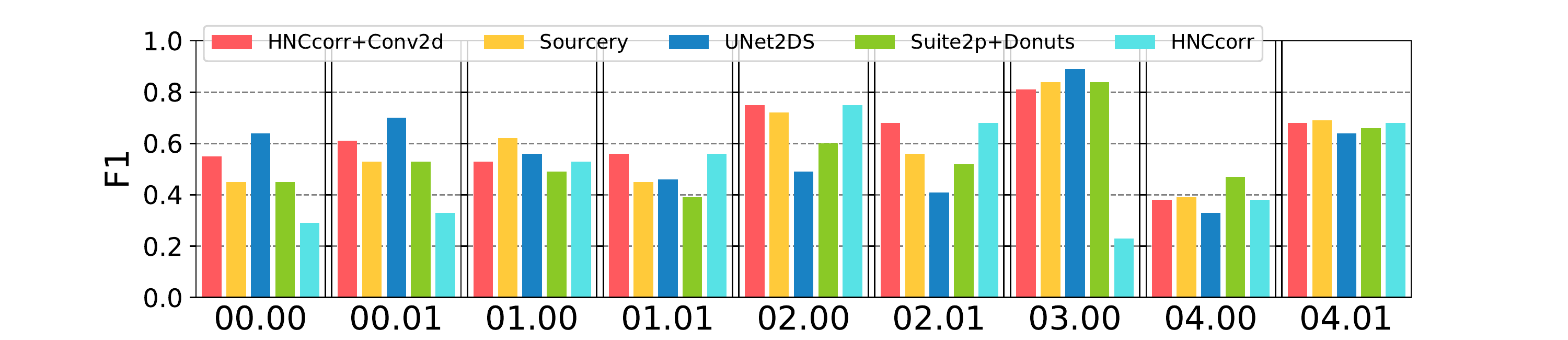}
    \caption{$F_1$ for the top five models on each of the Neurofinder testing datasets.}
    \label{fig:neurofinder_results}
\end{figure}

As of July 2017, the top-scoring Neurofinder submission is an ensemble of HNCcorr \cite{spaen2017hnccorr}, which clusters pixels based on correlation over time, and Conv2D \cite{conv2d}, a convolutional neural network with $40\times40$ input and $20\times20$ output windows. Second and fourth place use matrix factorization methods from \cite{pachitariu2013extracting, pachitariu2016suite2p}. U-Net2DS ranks third and is the highest-scoring model to exclusively use deep learning. Table \ref{tab:neurofinder_results} and Fig. \ref{fig:neurofinder_results} compare the top models' $F_1$, precision, and recall metrics. Models that explicitly use temporal information considerably outperform those using summaries on the 02.00 and 02.01 datasets. This might correspond to differences in ground truth labeling techniques acknowledged by the Neurofinder competition.\footnote{From the website: For the 00 data, labels are derived from an anatomical marker that indicates the precise location of each neuron and includes neurons with no activity. For the 01, 02, 03, 04 data, labels were hand drawn or manually curated, using the raw data and various summary statistics, some of which are biased towards active neurons.} To evaluate robustness beyond the Neurofinder datasets, we made predictions on the St. Jude datasets from the auditory cortex and had $F_1$ scores of 0.45 and 0.47 despite the datasets having been collected under different experimental settings and from a region of the brain distinct from all training datasets. Segmentations for two Neurofinder and two St. Jude datasets are shown in Fig. \ref{fig:neurofinder_predictions}.

\begin{table}[!t]
\begin{center}
\begin{tabularx}{\textwidth}{X|c|c|c}
\footnotesize

\textbf{Neurofinder Model} & \textbf{F$_1$} & \textbf{Precision} & \textbf{Recall} \\ \hline

HNCcorr \cite{spaen2017hnccorr} + Conv2d \cite{conv2d} & \hspace{0.1cm} $0.617 \pm0.122$ \hspace{0.1cm} & \hspace{0.1cm}$ 0.702 \pm0.170$ \hspace{0.1cm} & \hspace{0.1cm} $0.602 \pm0.197$ \hspace{0.1cm} \\ \hline
Sourcery \cite{sourcery} & \hspace{0.1cm} $0.583 \pm0.139$ \hspace{0.1cm} & \hspace{0.1cm}$ 0.599 \pm0.197$ \hspace{0.1cm} & \hspace{0.1cm} $0.629 \pm0.168$ \hspace{0.1cm} \\ \hline
UNet2DS & \hspace{0.1cm} $0.569 \pm0.160$ \hspace{0.1cm} & \hspace{0.1cm}$ 0.618 \pm0.235$ \hspace{0.1cm} & \hspace{0.1cm} $0.609 \pm0.185$ \hspace{0.1cm} \\ \hline
Suite2p \cite{pachitariu2016suite2p} + Donuts \cite{pachitariu2013extracting} & \hspace{0.1cm} $0.550 \pm0.127$ \hspace{0.1cm} & \hspace{0.1cm}$ 0.578 \pm0.156$ \hspace{0.1cm} & \hspace{0.1cm} $0.568 \pm0.181$ \hspace{0.1cm} \\ \hline
HNCcorr \cite{spaen2017hnccorr} & \hspace{0.1cm} $0.492 \pm0.180$ \hspace{0.1cm} & \hspace{0.1cm}$ 0.618 \pm0.169$ \hspace{0.1cm} & \hspace{0.1cm} $0.479 \pm0.268$ \hspace{0.1cm} \\ \hline

\end{tabularx}
\vspace{0.2cm}
\caption{Mean and standard deviation of metrics for the top five Neurofinder models.}
\vspace{-1cm}
\label{tab:neurofinder_results}
\end{center}
\end{table}

\subsubsection{Deep Learning Insights for Calcium Imaging Data}

We initially considered several pre-processing steps, including clipping maximum values to reduce noise and stretching each series' values to maximize contrast. These steps made it easier to recognize neurons visually but did not improve performance. The heavy neuron-to-background class-imbalance required attention in training. Some combinations of learning rates and loss functions caused the network to predict 100\% background early on and never recover. It helped to increase window size and monitor the proportion of neuron pixels predicted vs. the ground truth proportion. Finally, the training datasets have semantic differences (labeling technique, part of brain, etc.). In some cases we improved performance by training one model per dataset, but ultimately found it more practical to train a single model on all data.

\subsection{Other Deep Learning Models}

\subsubsection{U-Net2DS with Temporal Summaries}

By using mean summaries, U-Net2DS ignores temporal information and has difficulty separating overlapping neurons. One possible extension of U-Net2DS would add temporal inputs consisting of summary images along the time-axis (i.e., YZ, XZ summaries).

\subsubsection{Segmentation with Better 2D Summary Images} 

A summary image that amplifies neurons and minimizes background should improve a model like U-Net2DS. Taking inspiration from \cite{spaen2017hnccorr}, we experimented with summaries that compute the mean correlation and cosine distance of each pixel with its neighbors. Similar to \cite{pachitariu2016suite2p}, we found most neuron pixels had high similarity to nearby background pixels, making neighborhood similarity a poor summary function.

\subsubsection{Frame-by-frame Segmentation}

Summary images can preserve unwanted background noise, so it is plausible that a model trained to segment the series frame-by-frame could recognize a neuron in frames where surrounding noise is minimized. Unfortunately, it is difficult to select frames for training as most neurons blend with background at arbitrary times and only 2D labels are given. We tried a contrast-based weighting scheme that decreased loss for neurons with low contrast to surrounding background and increased loss for high-contrast neurons. Still, we found most neurons were more distinguishable in a mean summary than in their highest-contrast frames. It seems the summary smooths areas that are otherwise noisy in single frames.

\subsubsection{3D Convolutional Networks} 

Intuitively, directly including inputs from the time dimension should improve performance. We explored 3D convolutional networks, similar to \cite{milletari2016v, cciccek20163d, apthorpe2016automatic}.  Because series have variable length, we found it difficult to determine an input depth for all datasets. The 3D models were also prohibitively slow, taking at least an hour to segment a single series.

\section{Conclusion} 

Large labeled datasets like Neurofinder enable the use of deep learning for automated calcium imaging segmentation. We demonstrated U-Net2DS, an adaptation of U-Net \cite{ronneberger2015u} that uses mean summaries of calcium imaging series to quickly produce full segmentation masks while minimizing parameter adjustment, pre/post-processing, and assumptions for new datasets. Despite its relative simplicity, it is the best-performing non-ensemble and deep learning model in the Neurofinder competition. We described several other deep learning formulations  for this problem, and we believe a model that efficiently combines spatial and temporal information is a promising next step.

\subsubsection*{Acknowledgments}

This work was supported in part by the Department of Developmental Neurobiology at St. Jude Children's Research Hospital and by the U.S. Department of Energy, Office of Science, Office of Workforce Development for Teachers and Scientists (WDTS) under the Science Undergraduate Laboratory Internship program.

\bibliographystyle{splncs03}
\bibliography{refs}

\end{document}